\documentclass{article}

\usepackage[preprint]{nips_2018}

\usepackage[utf8]{inputenc}
\usepackage[T1]{fontenc}
\usepackage{hyperref}
\usepackage{url}
\usepackage{booktabs}
\usepackage{amsfonts}
\usepackage{nicefrac}
\usepackage{microtype}
\usepackage{array}

\usepackage{amssymb}
\usepackage{amsmath}
\usepackage{amsfonts}
\usepackage{graphicx}
\usepackage{subfigure}
\usepackage[normalem]{ulem}
\usepackage{refcount}
\usepackage{wrapfig}
\usepackage{gensymb}

\usepackage{color}
\definecolor{orange}{RGB}{235, 65, 65}
\definecolor{darkblue}{RGB}{0, 0, 83}

\definecolor{tan}{RGB}{113, 90, 58}

\definecolor{pink}{RGB}{226, 113, 187}

\newcommand{\model}[1]{{\small\textsc{#1}}}

\renewcommand{\cite}{\citep}
\bibpunct{(}{)}{;}{a}{}{,}

\title{Encoding Spatial Relations from Natural Language}

\author{
  Tiago Ramalho\thanks{Joint First authors.}, Tom\'a\v{s} Ko\v{c}isk\'y$^{*}$, \\
  \textbf{Frederic Besse, S. M. Ali Eslami, G\'abor Melis, Fabio Viola, Phil Blunsom,} \\
  \textbf{Karl Moritz Hermann} \\
  DeepMind, London, UK \\
  \texttt{\{tmramalho, tkocisky, kmh\}@google.com} \\
}

\begin{document}

\maketitle

\begin{abstract}
Natural language processing has made significant inroads into learning the semantics of words through distributional approaches, however representations learnt via these methods fail to capture certain kinds of information implicit in the real world. In particular, spatial relations are encoded in a way that is inconsistent with human spatial reasoning and lacking invariance to viewpoint changes. We present a system capable of capturing the semantics of spatial relations such as behind, left of, etc from natural language.
Our key contributions are a novel multi-modal objective based on generating images of scenes from their textual descriptions, and a new dataset on which to train it. We demonstrate that internal representations are robust to meaning preserving transformations of descriptions (paraphrase invariance), while viewpoint invariance is an emergent property of the system.
\end{abstract}

\section{Introduction}

Through natural language humans are able to evoke representations in each other's
minds. When one person describes their view of a scene their interlocutors are
able to form a mental model of the situation and imagine how the objects
described would look from different viewpoints. At the simplest level, if
someone in front of you describes an object as situated to their left, you
understand that it is to your right.
Current models for embedding the meaning of natural language are not able to achieve such viewpoint integration. In fact, as shown by \citet{Gershman2015}, distributed representations of natural language extracted from monolingual corpora fail to understand semantic equivalences such as that `A is in front of B' describes the same situation as `B is behind A.'

We believe an important first step toward a human-level ability to understand scene descriptions is building representations that can capture these invariances.
In this paper we introduce a multi-modal architecture that learns such representations.
In order to train and validate our model we create a large dataset of 3D scenes coupled with language descriptions from different viewpoints.
We focus on evaluating the learned representations to ensure that they are truly general, by generating images from angles not seen in the training data and checking whether they correspond to natural language descriptions of the same scene at the new angle. We also find that our learned representations align well with human similarity judgements of scene descriptions.

Spatial natural language is notoriously ambiguous and difficult to process
computationally \cite{Kranjec2014,Haun2011}. Even seemingly simple prepositions
like \textit{behind} are impossible to describe categorically and require a graded
treatment (e.g.\ how far may a person move from behind a tree before we no longer
describe them as such).
In addition, the lexicalisation of spatial concepts can vary widely across languages and cultures \cite{Haun2011}, with added complexity in how humans represent geometric properties when describing spatial experiences \cite{landau_jackendoff_1993} and in the layering of locatives \cite{Kracht2002}.
While there has been considerable research into the
relationship between categorical spatial relation processing, perception and
language understanding in humans, there are few definite conclusions on how to encode this relationship computationally \cite{Kosslyn1987,johnson1990body,Kosslyn1998,Haun2011}.

In the field of Natural Language Processing, work on spatial relations has
focused on the extraction of spatial descriptions from text and their mapping
into formal symbolic languages \cite{Kordjamshidi2012b,Kordjamshidi2012}, with
numerous such annotation schemes and methods proposed
\cite{Shen2009,Bateman2010,Rouhizadeh11}. Meanwhile research on
visualising spatial descriptions has predominantly employed heavily hand
engineered representations that do not offer the generic cross task advantages
of distributed representations \cite{Chang2014LearningSK,Hassani2016VisualizingNL}.

\section{Dataset of visually grounded scene descriptions}\label{sec:dataset}

\begin{figure}
\centering
\includegraphics[width=\textwidth]{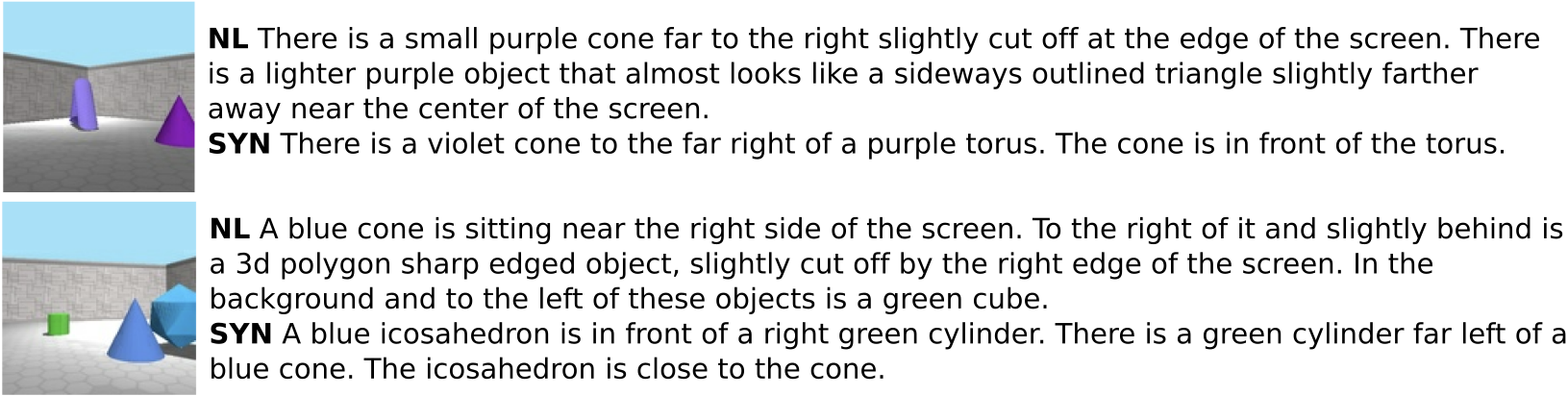}
\caption{%
Example descriptions with corresponding ground truth image. Shown are natural (NL) and
synthetic (SYN) language descriptions. Annotation linguistic errors have been preserved.
}
\label{fig:dataset}
\end{figure}

We construct virtual scenes with multiple views each presented in multiple modalities: image, and synthetic or natural language descriptions (see Figure~\ref{fig:dataset}). Each scene consists of two or three objects placed on a square walled room, and for each of the 10 camera viewpoint we render a 3D view of the scene as seen from that viewpoint as well as a synthetically generated description of the scene. For natural language we selected a random subset of these scenes and had humans describe a given scene and viewpoint based on the rendered image. Details on data generation are found in Appendix \ref{sec:appendix-data}.
We targeted a fairly low visual complexity of the rendered scenes since this factor is orthogonal to the spatial relationships we want to learn. The generated scenes still allow for large linguistic variety (see Table~\ref{tab:data}).
We publish the datasets described in this paper at \url{https://github.com/deepmind/slim-dataset}. For statistics see Table~\ref{tab:data}.
\begin{wraptable}[9]{r}{0.5\textwidth}
\caption{Dataset statistics.}
\label{tab:data}
\centering
\begin{tabular}{@{}lrr@{}}
\toprule
& Synthetic & Natural \\
\midrule
\# Training Scenes & 10M & 5,604 \\
\# Validation Scenes & 1M & 432 \\
\# Test Scenes & 1M & 568 \\
Vocabulary Size & 42 & 1,023 \\
Tokens per Description & 60 & 90\\
\bottomrule
\end{tabular}
\end{wraptable}

\textbf{Synthetic language data:}
\label{sec:syndata}
We generated a dataset of 10 million 3D scenes. Each scene contains two or three coloured 3D objects and light grey walls and floor. The  language descriptions are generated programmatically, taking into account the underlying scene graph and camera coordinates so as to describe the spatial arrangement of the objects as seen from each viewpoint.

\textbf{Natural language data:}
\label{sec:nldata}
We generated further scenes and used Amazon Mechanical Turk to collect natural language descriptions. We asked annotators to describe the room in an image as if they were describing the image to a friend who needs to draw the image without seeing it. We asked for a short or a few sentence description that describes object shapes, colours, relative positions, and relative sizes. We provided the list of object names together with only two examples of descriptions, to encourage diversity, while focusing the descriptions on the spatial relations of the objects.
The annotators annotated 6,604 scenes with 10 descriptions each, one for each view.

\section{Model description} \label{sec:model}

We propose a model that learns to integrate multiple descriptions of an underlying input into a single representation, and subsequently to generate new data from this representation in a multi-modal setup.
We refer to our model as the Spatial Language Integrating Model (SLIM).
This work is inspired by Generative Query Networks \cite{Eslami2018}, which integrate multiple visual inputs, and which can be used to generate new views of the same environment.
To put pressure on the representation to encode a viewpoint invariant scene description the model is setup so that it does not know which viewpoint will be decoded until after the representation is built.
In our case the model is given text descriptions of a scene as seen from $n$ different viewpoints to encode into a scene representation vector. This vector is then used to reconstruct an image of the scene as seen from a new viewpoint.

\begin{figure}
\centering
\includegraphics[width=0.75\textwidth]{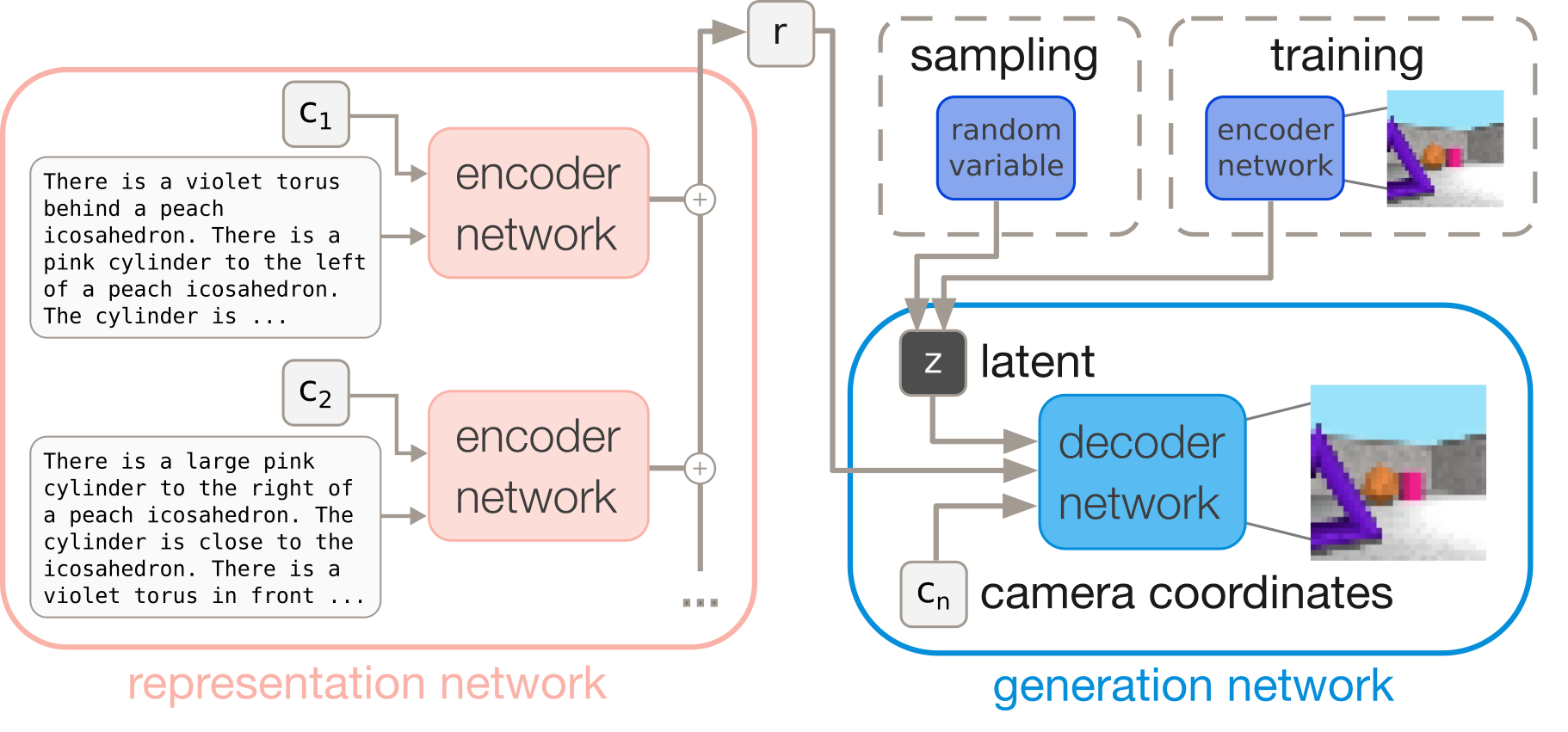}
\caption{Diagram of our model.
A representation network parses multiple descriptions of a scene from different viewpoints by taking in camera coordinates and a textual caption.
The representations for each viewpoint are aggregated into a scene representation vector $r$ which is then used by a generation network to reconstruct an image of the scene as seen from a new camera coordinate.
}
\label{fig:diagram}
\end{figure}

Our proposed model, depicted in Figure~\ref{fig:diagram}, consists of two parts:
a representation network that produces an aggregated representation from textual descriptions of the scene from a number of viewpoints, and a generation network conditioned on the scene representation that renders the scene as an image from a new viewpoint. We describe both networks below (see Appendix~\ref{sec:appendix-model} for minutiae).

The \textbf{representation network} encodes each text observation $d_i$ and corresponding camera angle $\theta_i$ into a viewpoint embedding $h_i$, ($i = 1\dots n$). The textual observation is encoded by a convolutional language encoder over the sequence of embedded words. This is concatenated with an embedding of the camera angle and passed through a multilayer perceptron  to obtain $h_i$.
We finally compute the scene representation
$r = \frac1n\sum_i^n h_i$,
independent of the number of inputs.
Clearly, more complex functions could be used as the aggregator, but for the purposes of the investigation presented here this relatively simple aggregation is sufficient.

The \textbf{generation network} is a conditional generative model that learns the distribution of likely images given a representation. To train the generative model, we sample an image $d_t$ and camera $c_t$ pair that was not provided to the representation network. This pair is used to train a conditional autoencoder \cite{NIPS2015_5775} where the conditioning variable is the concatenation of $r$ and $c_t$. Concretely, the autoencoder is composed of an encoder function which maps the image to a latent space $P(z) = e(d_t, c_t)$, and the decoder is a function which reconstructs the image from the a sample of the latent and the conditioning $\hat{d_t} = g(z, r, c_t)$.

We use a recurrent variational autoencoder, DRAW \cite{gregor2015draw}, to implement the generation network
and train the model by minimising the ELBO \cite{kingma2013auto} loss:
\begin{align*}
  \mathcal{L} = -\log D(x|r) + \sum_{k=1}^K KL \left(Q(z_k|h^{enc}_k)||P(z_k)\right)
\end{align*}
where $x$ is the image to be reconstructed. There are $K$ iterations in DRAW with $z_k$
denoting each latent for an iteration and $h^{enc}_k$ representing the encoded latent.
The reason for using the autoencoding component of the model is to guide the training, while simultaneously constraining it through the KL-term on an uninformative prior (in our case a zero mean unit variance gaussian).
While the autoencoder can help kick-start training, the model is strongly encouraged to decrease its reliance on $z$ and instead to extract all necessary information from $r$.

\section{Scene encoding experiments}\label{sec:experiments}

\begin{figure}
\centering
\includegraphics[width=\textwidth]{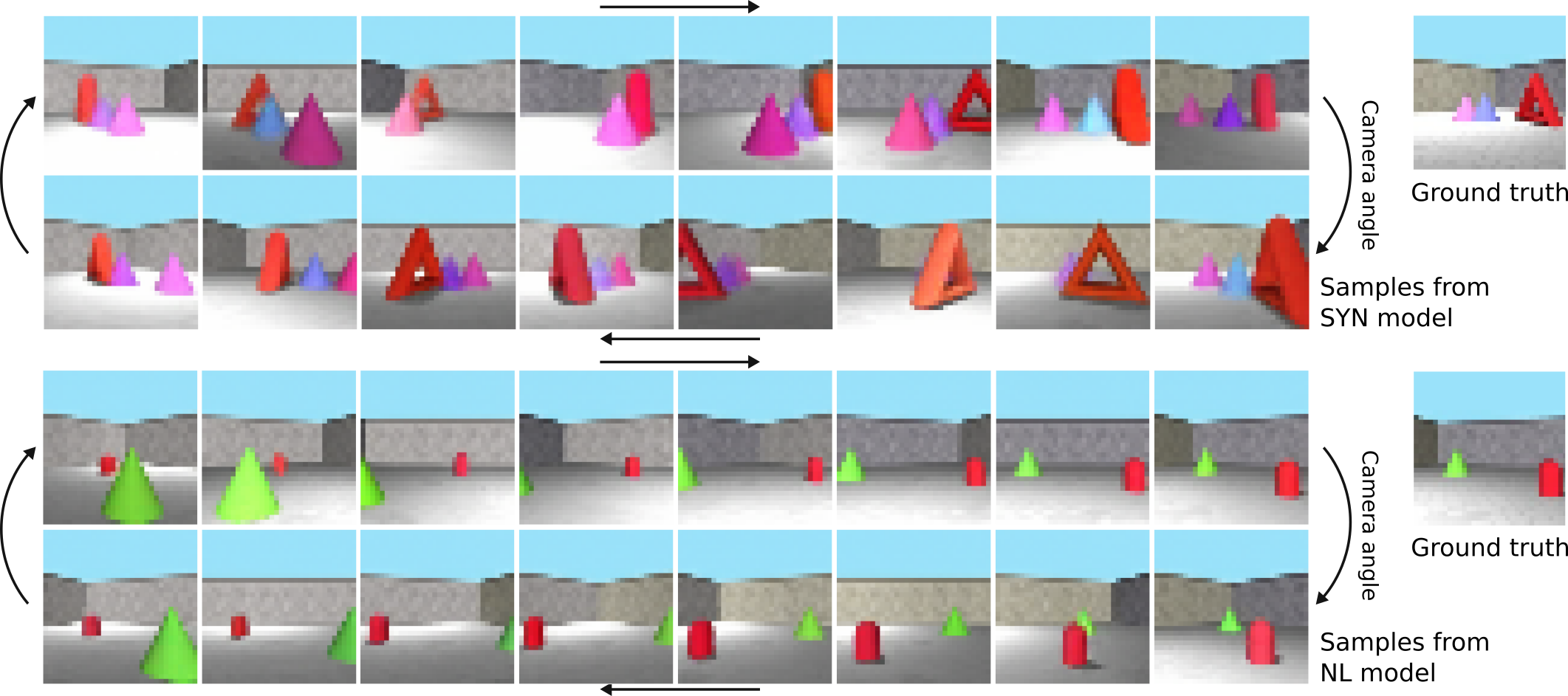}
\caption{Samples generated from the synthetic (top) and natural language (bottom) model.
Corresponding captions are: "There is a pink cone to the left of a red torus. There is a pink cone close to a purple cone. The cone is to the left of the cone. There is a red torus to the right of a purple cone."; "There are two objects in the image. In the back left corner is a light green cone, about half the height of the wall. On the right side of the image is a bright red capsule. It is about the same height as the cone, but it is more forward in the plane of the image."
}
\label{fig:samples}
\end{figure}

We first establish that our model learns to encode and integrate \emph{synthetic language} descriptions so that it can reconstruct images from new viewpoints.
Next, we evaluate the performance of the model on natural language scene descriptions in a number of
semi-supervised and domain adaptation training regimes, using synthetic data to support training the \emph{natural language} encoding model.

While pixel loss compared with gold images is an obvious metric to optimise against, we think that human judgements are more suitable here. We argue that pixel loss is a far too strict metric for evaluating the generative model due to the differing degrees of specificity between language and visual data points.
Consider the images and descriptions in Figure~\ref{fig:dataset}. Adjectives such as {\it big} or relative positions such as {\it to the far right} are sufficient to give us a high-level idea of a scene, but lack the precision required to generate an exact copy of a given image. The same description can be satisfied by an infinite number of visual renderings---implying that pixel loss with a gold image is not a precise measure of whether a given image is semantically consistent with the matching scene description.
Human judgements are a better way to evaluate this semantically and visually complex task of understanding spatial relations in scenes. In other words, we prefer spatial consistency over pixel-accuracy in the generated images.

We showed annotators model samples with corresponding language descriptions, and asked them to judge whether the descriptions matched, partially matched, or did not match the image. We additionally asked for binary choices on whether (1) all object shapes and (2) all object colours from the text are in the image; whether (3) all shape and colour combinations are correct; and whether (4) all objects are correctly positioned.
Annotators saw randomly mixed samples from all the models and benchmarks.
For the \model{Upper Bound} evaluation we showed gold images and matching descriptions.

Irrespective of training setup, we use synthetic descriptions of the output in the human annotation,
as these guarantee us a consistent degree of specificity versus natural language descriptions that can widely vary in that respect. Note that this only concerns the description shown to the human annotators. In the evaluation, we feed synthetic language into the model for the experiments in Section~\ref{sec:t2i}, and natural language descriptions in the case of Section~\ref{sec:sim2real}.

\subsection{Synthetic language experiments}\label{sec:t2i}

We trained SLIM on the synthetic text to verify whether it can be trained to understand scenes from descriptions. For each scene we randomly choose nine out of ten available views as inputs (text), and the final viewpoint as the target (image).
We trained with a batch size of 32 examples, using early stopping based on the validation data.

Figure~\ref{fig:samples} shows model samples and the underlying ground truth. We draw samples from the model by sampling the latents from the prior and varying the camera position.
Note that while the image samples exhibit a lot of variability and can differ widely from the ground truth image, they are consistent with the scene text description.

\begin{figure}
\centering
\includegraphics[width=\textwidth]{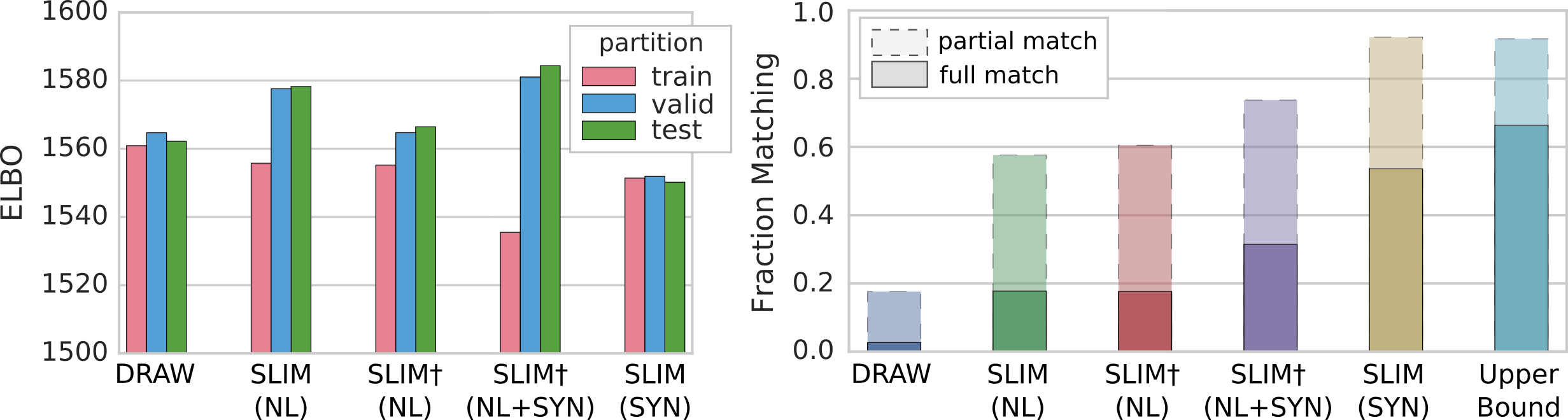}
\caption{ELBO numbers for the model variants under training for train/validation/test splits. Human ranking of consistency of visual scene samples with matching caption. For \model{SLIM${}^\dagger$ (NL+SYN)} numbers shown were calculated only on natural language inputs.}
\label{fig:elbo_human}
\end{figure}

We report ELBO loss on training, validation and test data in the left part of  Figure~\ref{fig:elbo_human}. As a baseline, we train an unconditional \model{DRAW} generative model (i.e.\ no scene descriptions are provided) which learns the unconditional output image distribution.
The ELBO captures how well the model can model the distribution of the images of the scene, and the results suggest that all setups under consideration are capable of doing this. However the ELBO does not capture the conditional likelihood given the representation and therefore does not show a significant difference between our conditional setup and the unconditional model.

Figure~\ref{fig:elbo_human} right shows the results of the human evaluation. Naturally, samples from the unconditional baseline perform poorly as it does not condition on the descriptions used as basis for the human judgements. However, these random samples still serve a purpose in informing us about the complexity of the dataset as well providing a lower bound on the generosity of the human annotators, so to speak.

The \model{Upper Bound} results are from annotators comparing ground truth images with their matching synthetic description. Note here that with 66.39\% perfect matches and 91.70\% including partial matches, these results are far from perfect. This indicates several issues: annotators may have used a very strict definition of a perfect match, penalising the fact that the synthetic descriptions are succinct and leaving out details such as the background colours for instance. Gold standard synthetic captions can also be perceived as incorrect due to factors such as the description of relative locations\footnote{Consider two objects behind each other with one shifted slightly to the left of the other. Their relationship could be described as "behind and left", just "behind" and under some circumstances also as just "to the left of", with different annotators preferring different schemes and considering others invalid.} or colour names not matching an annotators understanding.
Our model, \model{SLIM (SYN)}, matches the performance of the \model{Upper Bound} in the human judgements, scoring 92.19\% including partial matches, while underperforming slightly with respect to perfect matches.

\subsection{Natural language experiments}
\label{sec:sim2real}

Following the synthetic language experiments, we investigate whether SLIM can cope with natural language.
Due to cost of annotation, the amount of natural language training data is several orders of magnitude smaller than the synthetic data used in the previous section (see Table~\ref{tab:data}).
We consider multiple training regimes that aim to mitigate the risk of over-fitting to this small amount of data.
The na\"{i}ve setup (\model{SLIM (NL)} in Figure \ref{fig:elbo_human}) uses only the natural language training data.
We also took a frozen pre-trained generation network (trained on synthetic data) and trained the representation network alone on the same dataset (\model{SLIM${}^\dagger$ (NL)}). Finally, \model{SLIM${}^\dagger$ (NL+SYN)} uses the same pre-training and uses an augmented dataset with a 50/50 split between synthetic and natural language.
We evaluate all three models with natural language inputs only, independent of training protocol.

We follow the training regime described above for the synthetic data. In addition, we found that adding 50\% dropout together with early stopping on validation data provided the best performance.
The drastic increase in complexity compared with the synthetic task is evident in the lower scores.
However, and more importantly, the relative performance of the three natural language training setups is consistent across all training runs, with the joint model (frozen decoder, encoder trained jointly on synthetic and natural language) performing best and the model trained on natural language only performing the worst (Figure~\ref{fig:elbo_human}).
This is encouraging, both as the absolute numbers suggest that the SLIM architecture is capable of encoding spatial relations from natural language, and as the significant gains in the joint training regime over the na\"{i}ve approach highlight the potential of using this type of data for further research into simulation to real world model adaptation.

\section{Representation analysis}\label{sec:analysis}

\begin{figure}
\centering
\includegraphics[width=\textwidth]{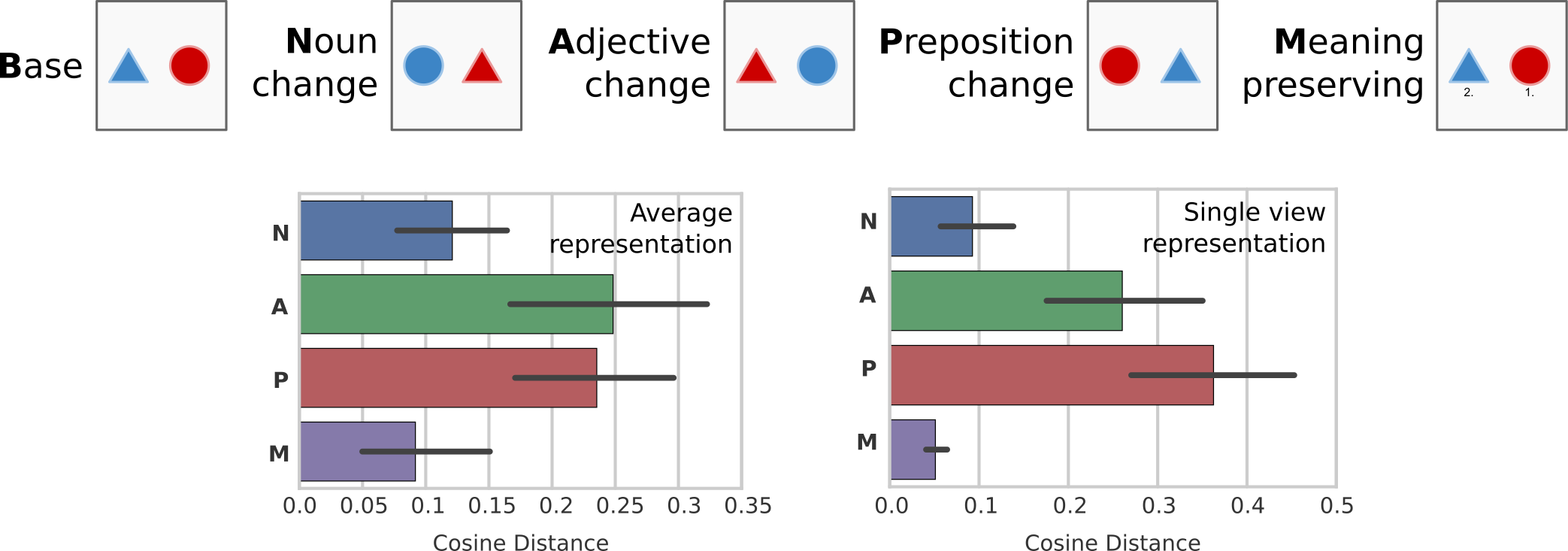}
\caption{Top, visual diagram of scene transformations used to test representational similarity. Left bottom, cosine distance between base representation and representation induced by applying one of the four transformations to each of the context inputs. Right bottom, same analysis, but for the representation generated by a single encoder step. Black bars represent 95\% CI. For comparison, the average human ranking would be $\mathrm{M > P > A > N}$ \citep{Gershman2015}.}
\label{fig:rep_sim}
\end{figure}

Having established that SLIM can successfully create a representation of the scene to reconstruct an image view, we now turn our attention to how this representation is built up from language descriptions of the scene.
We show that our model does learn semantically consistent representations, unlike the neural network models investigated in \citet{Gershman2015}.
Secondly, we examine representations at different stages of our model to establish at what point this semantic coherence arises, and similarly, at what point viewpoint invariance (independence of input camera angles) is achieved, allowing us to recreate scenes from new viewpoints. The following set of equivalences exemplifies the difference between the two concepts:

\begin{table}[h!]\centering
\vspace{-0.6em}
\begin{tabular}{@{}llll@{}}
\toprule
{\bf Paraphrase invariance} & (A left of B) &$\equiv$ &(B right of A) \\
{\bf Viewpoint invariance} &  (A left of B, $0\degree$) &$\equiv$ &(A right of B, $180\degree$)\\
\bottomrule
\end{tabular}
\vspace{-0.6em}
\end{table}

\subsection{Semantic coherence analysis}
\label{sec:transformations}

\citet{Gershman2015} consider sentence transformations (Table~\ref{tab:tenenbaum}) and compare human similarity judgements of these transformations, to the base sentence, with the distances of the output of neural network models.
That work showed that
recurrent language encoding models fail to coherently capture
these semantic judgements in output representations.
\begin{table}[t]\centering
\caption{\citet{Gershman2015} transformations.}
\label{tab:tenenbaum}
\begin{tabular}{@{}lll@{}}
\toprule
{\bf B} &Base sentence & "a young woman in front of an old man." \\
{\bf N} &Noun change & "a young man in front of an old woman." \\
{\bf A} &Adjective change & "an old woman in front of a young man." \\
{\bf P} &Preposition change & "a young woman behind an old man." \\
{\bf M} &Meaning preservation & "an old man behind a young woman." \\
\bottomrule
\end{tabular}
\end{table}

Human annotators \emph{rank} the transformations of the base sentence, B, in order of semantic similarity, resulting in average rankings $\mathrm{M>P>A>N}$, from most to least similar. In \citet{Gershman2015} a set of neural language encoders is then shown not to follow this human judgement. In particular, the M transformation---semantically equivalent---is never the most similar to the base sentence.
We believe that the original experiment's conclusions may have been too broad: first, as the authors themselves state, the pre-trained language models considered were built on English Wikipedia and the Reuters RCV1 corpus \cite{Collobert2011}, neither of which exhibit significant amounts of language in the style of the test data. Next, there was nothing in the objective functions for these models that would encourage solving the task at hand. This is the key question: is the deflationary criticism with respect to the model architectures valid, or is the failure to solve this task a function of the training regime?

We posit that a learning objective grounded in another modality---such as the visual domain---could learn spatial semantic coherence by directly teaching the model a complex function to reconcile syntax (the words) and semantics (the meaning, here captured by the scene and its visual representation).
We test this hypothesis by re-implementing the sentence transformation setup in our multi-modal setup.

We have four scene templates, as depicted on Figure~\ref{fig:rep_sim}, matching the transformations in \citet{Gershman2015}, and generate synthetic descriptions analogous to the examples in Table~\ref{tab:tenenbaum} from these. The meaning preserving transformation, M, naturally results in the same scene as the base scene; however the resultant descriptions match the original transformation from B to M.

For this analysis we use a model trained on synthetic data (see Section \ref{sec:t2i}). We separately encode the set of descriptions for base scene ($r$) and transformations ($r_n,r_a,r_p,r_m$) and compute their similarity using the cosine distance between the representations%
.\footnote{Euclidean distance and Pearson's r result in the same ordering of similarities.}
The results are depicted in Figure~\ref{fig:rep_sim}, averaged over 3,200 scenes and their transformations. Unlike the results in \citet{Gershman2015}, the SLIM architecture encodes B and M as the most similar, both when considering a single sentence encoding and even when considering the full scene representations.

Of the semantically different descriptions, the noun transformations produce the least dissimilar representations. We believe this is due to the fact that the reconstruction loss is based on a pixel measure, where a colour change (adjective change) leads to a larger loss in terms of the overall number of pixels changed in the visual representation than the noun change which only alters the shape of the objects, causing a smaller number of pixels to change in the overall visual representation.

This reinforces our earlier point that in order to properly encode the kind of syntactic/semantic divergence tested for by this analysis, we need models that train not only on syntax but also on semantics. While the visual domain serves as a useful proxy for semantics in our setup, the pixel loss analysis highlights that it is only a proxy with its own shortcomings and biases.

\subsection{Encoder hierarchy representation analysis}

Here we show that while individual encodings are strongly view-dependent, the aggregation step and decoder integrate these representations into a viewpoint invariant end-to-end model.
We investigate how representations for the same scene differ
at a number of stages: after encoding the language and camera angle, after the aggregation step, and finally after decoding.

\begin{figure}
\includegraphics[width=\columnwidth]{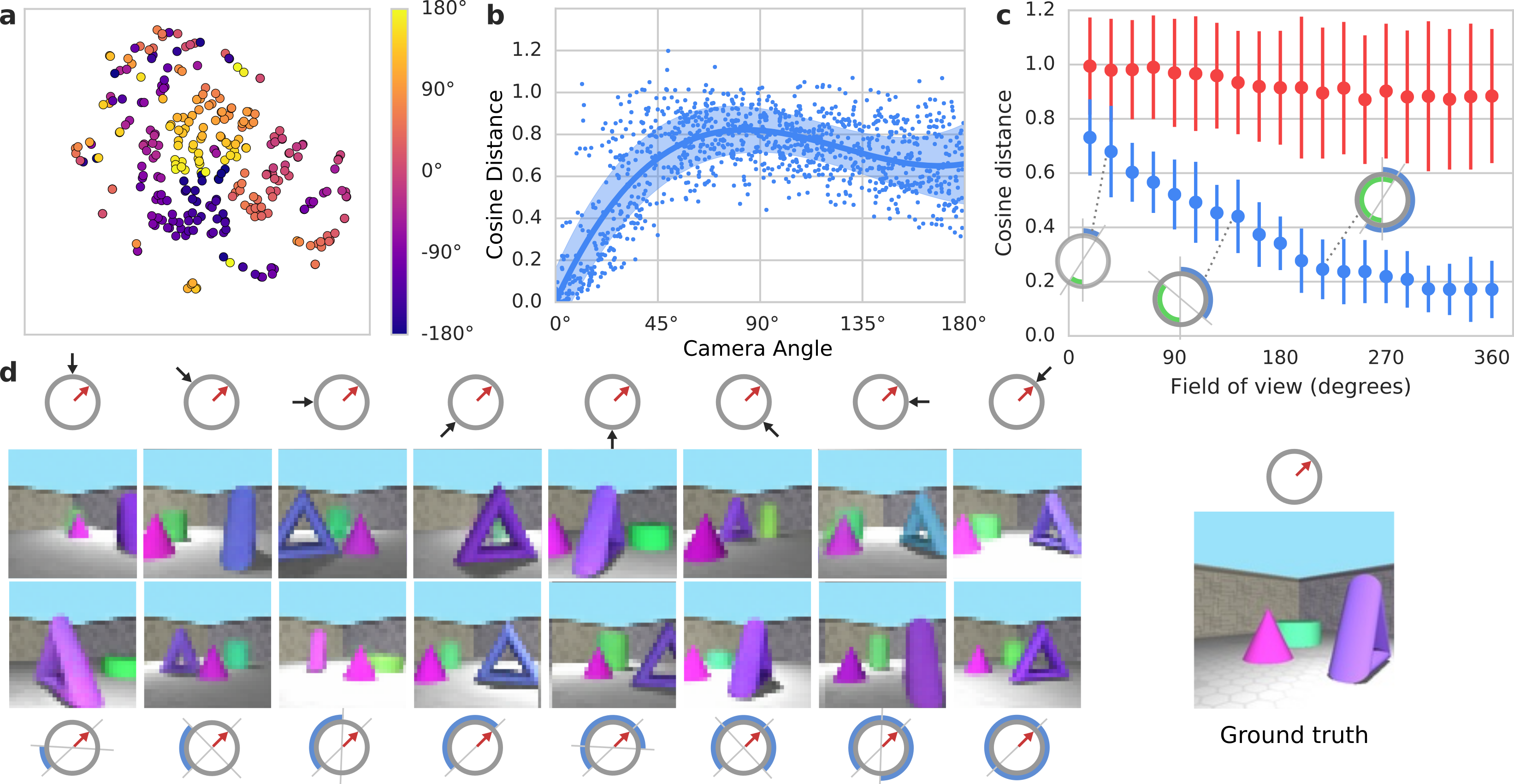}
\caption{%
a) t-SNE of single description encodings, coloured by camera angle.
b) Distance between single description representations of the same scene as a function of the angle between their viewpoints.
c) Distance between aggregated representations drawn from opposing arcs as a function of the size of those arcs. Blue compares same scene representations, red different scene representations.
d) Output samples for a constant scene and coordinate under varying input conditions. Top: a single description (from black arrow), bottom: aggregated descriptions from an increasingly sized arc.}
\label{fig:scatters}
\end{figure}

Figure~\ref{fig:scatters}a is a t-SNE plot of the intermediate representations for a number of scenes. Each point in that plot corresponds to a single language and coordinate embedding, with embeddings for all 10 views in 32 scenes shown. The embeddings cluster by camera angle even though they are sampled from different scenes. This shows how on a superficial level the camera coordinates dominate the single input representations.
Figure~\ref{fig:scatters}b reinforces this analysis: If a camera independent representation were built at the language encoder level, we would expect each intermediate representations to be similar,
with the aggregation process only reducing noise. The quickly diverging cosine distances as a function of the angle between two camera coordinates shows that this is not the case, and it is clear that the representations do not have the desired property at this stage in the model.

Next we investigate the effect of the aggregation function. Consider a circle around a scene, with arcs originating at $0\degree$ and $180\degree$, respectively. We clockwise increase the central angle $\theta$ of both arcs simultaneously and calculate aggregated representations of nine language descriptions from within each arc.\footnote{Sampled with replacement.} Figure~\ref{fig:scatters}c (blue) shows how the cosine distance between these representations decreases as the size of the two arcs increases. Note that up while $\theta\le\pi$, the two representations are non-overlapping, demonstrating how even the simple additive aggregation function is able to cancel out multiple viewpoints while integrating their information for added robustness. If the representations corresponding to each arc are sampled from two different scenes (red), the representations no longer converge as more data is integrated.
Together these analyses suggest that the aggregation step is doing something beyond cancelling out noise such as cancelling out the non-invariant parts of the encoding.

Figure \ref{fig:scatters}d shows that the full model achieves viewpoint invariance. We sample images from the model using the same output coordinates and input scene, but vary what information we input. Regardless of the number of input descriptions or their location, we receive images that are broadly consistent with the semantics of the gold scene. As the input gets closer to the target or as the number of viewpoints sampled increases, the samples become semantically more consistent with the ground truth.

\section{Conclusion}

We have presented a novel architecture that allows us to represent scenes from language descriptions. Our model captures both paraphrase and viewpoint invariance, as shown through a number of experiments and manual analyses.
Moreover, we demonstrated that the model can integrate these language representations of scenes to reconstruct a reasonable image of the scene, as judged by human annotators.
Lastly, we demonstrated how to effectively use synthetic data to improve the performance of our model on natural language descriptions using a domain adaptation training regime.

This paper serves to highlight one key point, namely the importance of aligning training paradigms with the information one wishes to encode. We demonstrated how such information can be provided by merging multiple modalities.
More broadly, this also demonstrates that the criticism in \citet{Gershman2015} is not an in-principle problem of the model, but rather a matter of setting up your training objectives to match your desired outcomes.

\bibliography{bibliography}
\bibliographystyle{plainnat}

\newpage
\appendix

\section{Model details}\label{sec:appendix-model}

\subsection{Representation network}
The representation network is composed of an encoder and an aggregation step. The input to representation network is a sequence of pairs $(d_i, \theta_i)$. The encoder network consists of a convolutional language encoder over the sequence of embedded words (embedding dimension 64) $\hat{d}_i = \mathrm{CNN}\!\left(\text{embed}(d_i)\right)$. In particular, we use a variant of ByteNet \cite{kalchbrenner2016neural} with three 1D convolutional layers,  dilation factor of 2 and layer normalisation followed by average aggregation over the sequence dimension.
The camera is represented by the embedding
$\hat{c}_i=\mathrm{MLP} ([\cos(\theta_i), \sin(\theta_i)])$ (dimension 32).
Next, these two representations are merged via a three-layer residual MLP to generate an embedding vector
$h_i= \mathrm{MLP} (\hat{d}_i \| \hat{c}_i )$
with dimensionality 256.
The aggregation step consists of computing the scene representation
$r = \frac1n\sum_i^n h_i$,
independent of the number of inputs. For training $n=9$.

\subsection{Generation network}

To train the generative model, we sample a target pair ($d_t$, $\theta_t$) that was not provided to the representation network. This pair is used to train a conditional autoencoder \cite{NIPS2015_5775} where the conditioning variable is the concatenation of $r$ and $\theta_t$.

We use the DRAW \cite{gregor2015draw} network to implement the generation network. DRAW is a recurrent variational autoencoder which has been used successfully to render images of complex scenes such as the 3D images in the dataset. We use a DRAW model with 12 iterations and with a convolutional LSTM core with dimensionality 128. The conditioning variable is concatenated with the sampled latent $z$ at each iteration. The output distribution is a Bernoulli on each subpixel. We train the model by minimizing the ELBO \cite{kingma2013auto}, which is made up of the total likelihood of the target image under this distribution plus the KL for the DRAW prior:
\begin{align*}
   \mathcal{L} = -\log D(x|r) + \sum_{k=1}^K KL \left(Q(z_k|h^{enc}_k)||P(z_k)\right)
\end{align*}
where $x$ is the image to be reconstructed and $r$ is the representation created by the encoders. The KL is the sum of terms from each of the $K$ iterations in draw with $z_k$ denoting each latent for an iteration and $h^{enc}_k$ representing the encoded latent.

\subsection{Training details}

We train our model using the ADAM optimizer with a learning rate annealing schedule starting at $5e-4$ and decaying linearly to $5e-5$ over one million steps. Training is stopped at the minimum validation loss calculated every 500 steps for 3200 samples (100 minibatches). For the synthetic dataset we use dropout of 0\% while for natural language we use 50\% dropout.

\section{Dataset generation details}\label{sec:appendix-data}

To generate the synthetic scenes we employed the MuJoCo engine\footnote{\url{http://www.mujoco.org/}} to create a generic 3D square room where multiple simple geometric objects are placed at random. The room is represented as a square with coordinates $x, y \in [-1, 1]$, with object coordinates sampled uniformly. Each scene contains two or three objects the identities of which are determined by a set of three latent variables: 8 shapes\footnote{%
cube, box, cone, triangle, cylinder, capsule, icosahedron, sphere\label{footnote:objects}}; a HSV color sampled uniformly in the intervals $H \in [0,1];\;S\in[0.5, 1];\;V\in[0.8, 1]$; and object size
\footnote{a scaling factor for the object mesh chosen uniformly from an interval sensible for the MuJoCo renderer}.
Lastly, we randomly sample camera positions from a circle centred at the middle of the room, with a radius approximately equal to the distance from the walls to the center.

To create a data batch we sample $10$ camera positions and for each we render a view of the scene from that angle and associate with it a textual description. The generated images are $128\times 128$ RGB images which are downsampled to $32\times 32$ and rescaled to floating point values in the range $[0,1]$ before being fed to the model.

The first part of our dataset contains synthetic language descriptions where the descriptions are generated by a script with access to the scene geometry. We iterate over object pairs, and for each object determine its shape name (matching the shapes described above, a color name (found by looking up the nearest HSV neighbor in a look-up table with 22 named colours) and a size name (large, small or no size). A description is then sampled by generating a caption relating those two objects with a spatial relation (in front of, behind, left of, right of, close, far). The order of objects in the description is randomly sampled. The final caption is composed of enough such descriptions so as to cover every pairwise relation for all objects in the scene exactly once.

To generate the natural language dataset we subsampled scenes from the synthetic dataset and asked humans to write a description of the rendered image of the scene from a certain camera position.

We have created validation and test sets based on held out combinations of colour and object type for each of the synthetic and natural language labeled scenes, with different combinations held out in the validation and the test data, and a guarantee that each validation/test scene contains at least one object unobserved during training.

The training set does not contain objects from the set \{`yellow sphere', `aqua icosahedron', `mint torus', `green box', `pink cylinder', `blue capsule', `peach cone'\}. The validation set scenes contain at least one of the first three elements of the held out combinations set and none of the remaining four elements. The test set scenes can contain any combination of shapes or colours, making sure they contain at least one of the last four combinations in the held out set.

\subsection{Dataset examples}

\subsubsection{Synthetic language, two objects}

\begin{tabular}{@{}m{2.2cm}m{11cm}@{}}
\toprule
\includegraphics[trim={0 13.6cm 54.4cm 0},clip, width=2.0cm]{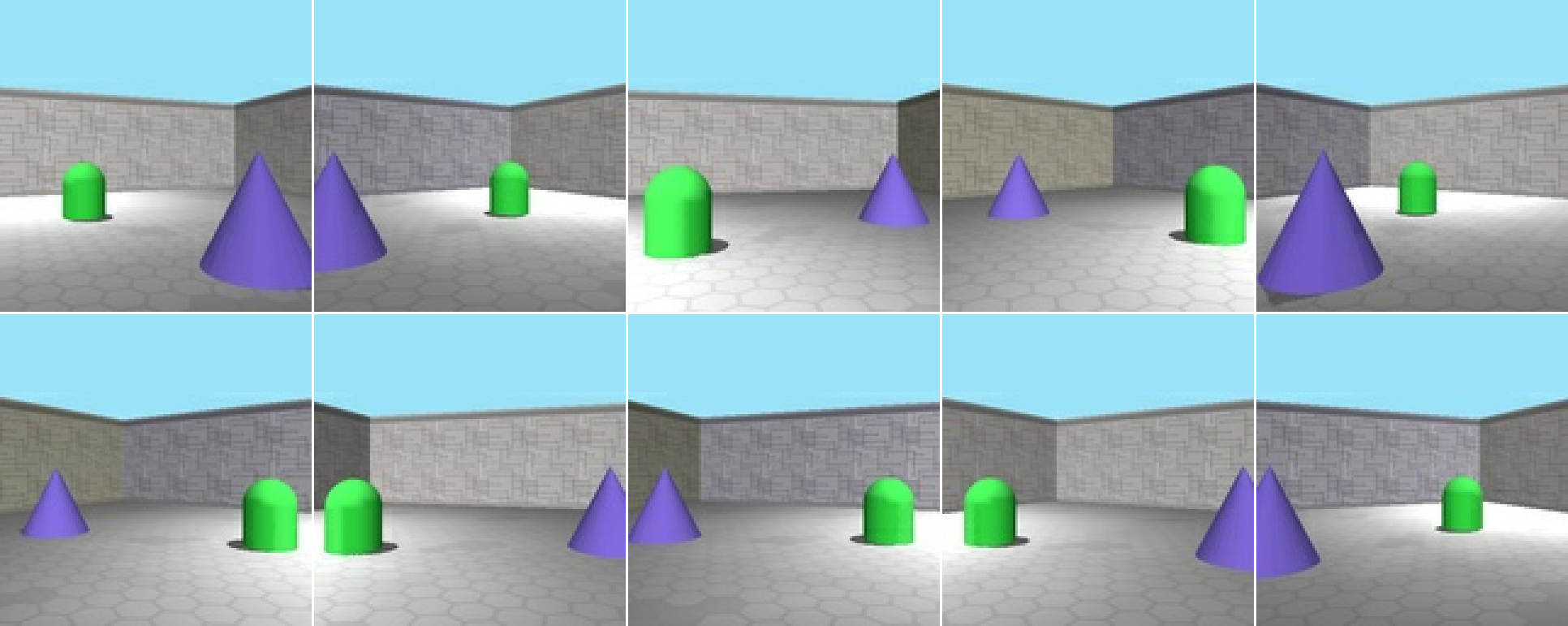} & There is a green capsule behind a purple cone. The capsule is
to the left of the cone.\\
\includegraphics[trim={13.6cm 13.6cm 40.8cm 0},clip, width=2.0cm]{syn_2obj} & There is a purple cone to the left of a green capsule. The cone
is in front of the capsule.\\
\includegraphics[trim={27.2cm 13.6cm 27.2cm 0},clip, width=2.0cm]{syn_2obj} & There is a green capsule in front of a purple cone. The capsule
is to the left of the cone.\\
\includegraphics[trim={40.8cm 13.6cm 13.6cm 0},clip, width=2.0cm]{syn_2obj} & There is a green capsule to the right of a purple cone. The
capsule is in front of the cone.\\
\includegraphics[trim={54.4cm 13.6cm 0 0},clip, width=2.0cm]{syn_2obj} & There is a purple cone in front of a green capsule. The cone is
to the left of the capsule.\\
\includegraphics[trim={0 0 54.4cm 13.6cm},clip, width=2.0cm]{syn_2obj} & There is a green capsule to the right of a purple cone. The
capsule is in front of the cone.\\
\includegraphics[trim={13.6cm 0 40.8cm 13.6cm},clip, width=2.0cm]{syn_2obj} & There is a green capsule in front of a purple cone. The capsule
is to the left of the cone.\\
\includegraphics[trim={27.2cm 0 27.2cm 13.6cm},clip, width=2.0cm]{syn_2obj} & There is a green capsule in front of a purple cone. The capsule
is to the right of the cone.\\
\includegraphics[trim={40.8cm 0 13.6cm 13.6cm},clip, width=2.0cm]{syn_2obj} & There is a green capsule to the left of a purple cone. The
capsule is behind the cone.\\
\includegraphics[trim={54.4cm 0 0 13.6cm},clip, width=2.0cm]{syn_2obj} & There is a green capsule behind a purple cone. The capsule is
behind the cone.\\
\bottomrule
\end{tabular}

\subsubsection{Synthetic language, three objects}

\begin{tabular}{@{}m{2.2cm}m{11cm}@{}}
\toprule
\includegraphics[trim={0 13.6cm 54.4cm 0},clip, width=2.0cm]{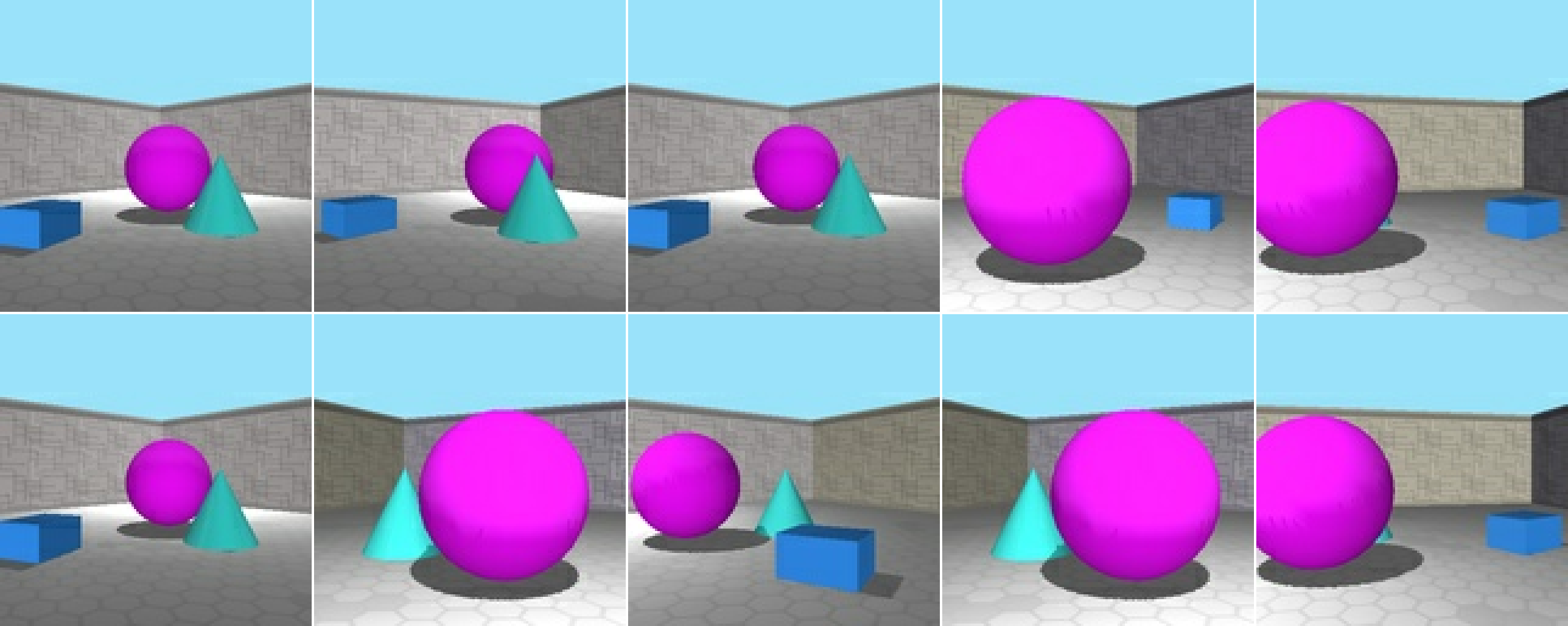} & There is a large magenta sphere behind a aqua cone. The sphere
is behind the cone. There is a aqua cone to the right of a blue box.
There is a blue box in front of a large magenta sphere.\\
\includegraphics[trim={13.6cm 13.6cm 40.8cm 0},clip, width=2.0cm]{syn_3obj} & There is a aqua cone in front of a large magenta sphere. There
is a aqua cone to the right of a blue box. There is a large magenta
sphere to the right of a blue box.\\
\includegraphics[trim={27.2cm 13.6cm 27.2cm 0},clip, width=2.0cm]{syn_3obj} & There is a large magenta sphere behind a aqua cone. The sphere
is behind the cone. There is a aqua cone to the right of a blue box.
There is a large magenta sphere behind a blue box. The sphere is
behind the box.\\
\includegraphics[trim={40.8cm 13.6cm 13.6cm 0},clip, width=2.0cm]{syn_3obj} & There is a aqua cone behind a large magenta sphere. The cone is
behind the sphere. There is a aqua cone to the left of a blue box.
There is a blue box to the right of a large magenta sphere.\\
\includegraphics[trim={54.4cm 13.6cm 0 0},clip, width=2.0cm]{syn_3obj} & There is a large magenta sphere in front of a aqua cone. There
is a aqua cone to the left of a blue box. There is a large magenta
sphere to the left of a blue box.\\
\includegraphics[trim={0 0 54.4cm 13.6cm},clip, width=2.0cm]{syn_3obj} & There is a aqua cone in front of a large magenta sphere. There
is a aqua cone to the right of a blue box. There is a large magenta
sphere behind a blue box. The sphere is behind the box.\\
\includegraphics[trim={13.6cm 0 40.8cm 13.6cm},clip, width=2.0cm]{syn_3obj} & There is a aqua cone to the left of a large magenta sphere.
There is a blue box to the right of a aqua cone. There is a blue box
behind a large magenta sphere. The box is behind the sphere.\\
\includegraphics[trim={27.2cm 0 27.2cm 13.6cm},clip, width=2.0cm]{syn_3obj} & There is a aqua cone to the right of a large magenta sphere.
There is a blue box in front of a aqua cone. There is a large magenta
sphere to the left of a blue box.\\
\includegraphics[trim={40.8cm 0 13.6cm 13.6cm},clip, width=2.0cm]{syn_3obj} & There is a aqua cone to the left of a large magenta sphere.
There is a aqua cone to the left of a blue box. There is a blue box
behind a large magenta sphere. The box is behind the sphere.\\
\includegraphics[trim={54.4cm 0 0 13.6cm},clip, width=2.0cm]{syn_3obj} & There is a aqua cone behind a large magenta sphere. The cone is
behind the sphere. There is a blue box to the right of a aqua cone.
There is a large magenta sphere to the left of a blue box.\\
\bottomrule
\end{tabular}

\subsubsection{Natural language, two objects}

\begin{tabular}{@{}m{2.2cm}m{11cm}@{}}
\toprule
\includegraphics[trim={0 13.6cm 54.4cm 0},clip, width=2.0cm]{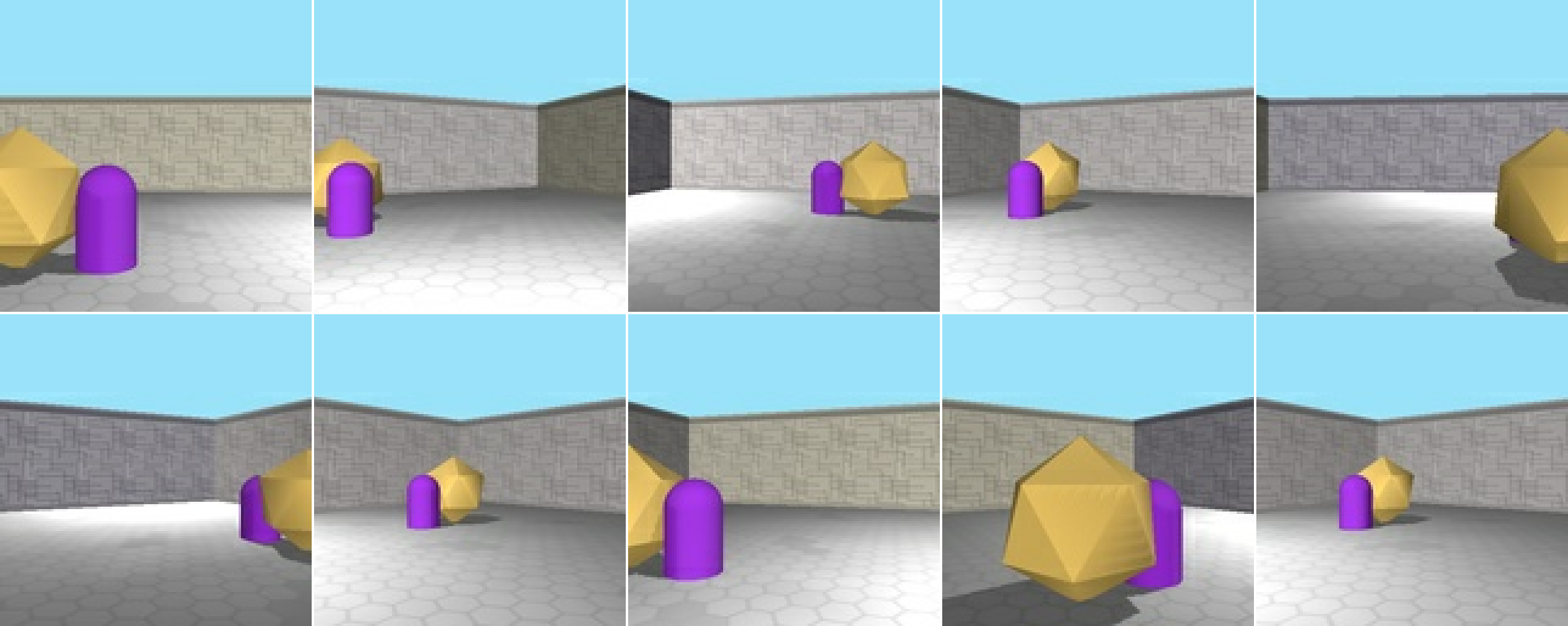} & a room has two grey walls and a light blue ceiling. a large
yellow cylinder is on the left next to a pink tube in the center.\\
\includegraphics[trim={13.6cm 13.6cm 40.8cm 0},clip, width=2.0cm]{nl_2obj} & Corner of a room with a blue ceiling grey brick walls and grey
tile floor. There are two objects in the room and the are on the left
side of the room furthest from the corner. Object one is a medium size
purple dome and it is in front of the larger medium sized gold double
pointed round cylinder. The room is three dimensional as well as the
objects.\\
\includegraphics[trim={27.2cm 13.6cm 27.2cm 0},clip, width=2.0cm]{nl_2obj} & in grey room, under a blue sky a tan ball sits next to a purple
rounded cylinder against a grey brick wall.\\
\includegraphics[trim={40.8cm 13.6cm 13.6cm 0},clip, width=2.0cm]{nl_2obj} & a room has two grey walls and a light blue ceiling. a yellow
ball is in the center.\\
\includegraphics[trim={54.4cm 13.6cm 0 0},clip, width=2.0cm]{nl_2obj} & In this image a gold dodecagon is in the right corner.\\
\includegraphics[trim={0 0 54.4cm 13.6cm},clip, width=2.0cm]{nl_2obj} & On the very right side of the room, in the center, is a tan
prism shape with the bottom resembling a sphere, only the left half is
visible. It is roughly half the size of the wall behind it. Behind
that shape is a purple rod shape that is shorter. The right side is
partially obscured by the shape in front of it.\\
\includegraphics[trim={13.6cm 0 40.8cm 13.6cm},clip, width=2.0cm]{nl_2obj} & A purple shape that is like the end of a hotdog. Shape with 3
triangles on top and bottom to form points and 8 triangles that round
out the middle\\
\includegraphics[trim={27.2cm 0 27.2cm 13.6cm},clip, width=2.0cm]{nl_2obj} & To the far left is a small purple bullet with a tan 10 sided
shape to the left border.\\
\includegraphics[trim={40.8cm 0 13.6cm 13.6cm},clip, width=2.0cm]{nl_2obj} & A yellow decagon sits in front of a purple cylinder which ends
with a dome. Walls of grey brick converge behind them. Floor is grey
octagons.\\
\includegraphics[trim={54.4cm 0 0 13.6cm},clip, width=2.0cm]{nl_2obj} & Corner of a room, blue ceiling with grey brick walls and a grey
tiled floor, There are two objects in the room and towards the back
corner. Object one is a purple slender dome and behind it slightly to
the right is a gold hexagon. Both objects are 3 dimensional.\\
\bottomrule
\end{tabular}

\subsubsection{Natural language, three objects}

\begin{tabular}{@{}m{2.2cm}m{11cm}@{}}
\toprule
\includegraphics[trim={0 13.6cm 54.4cm 0},clip, width=2.0cm]{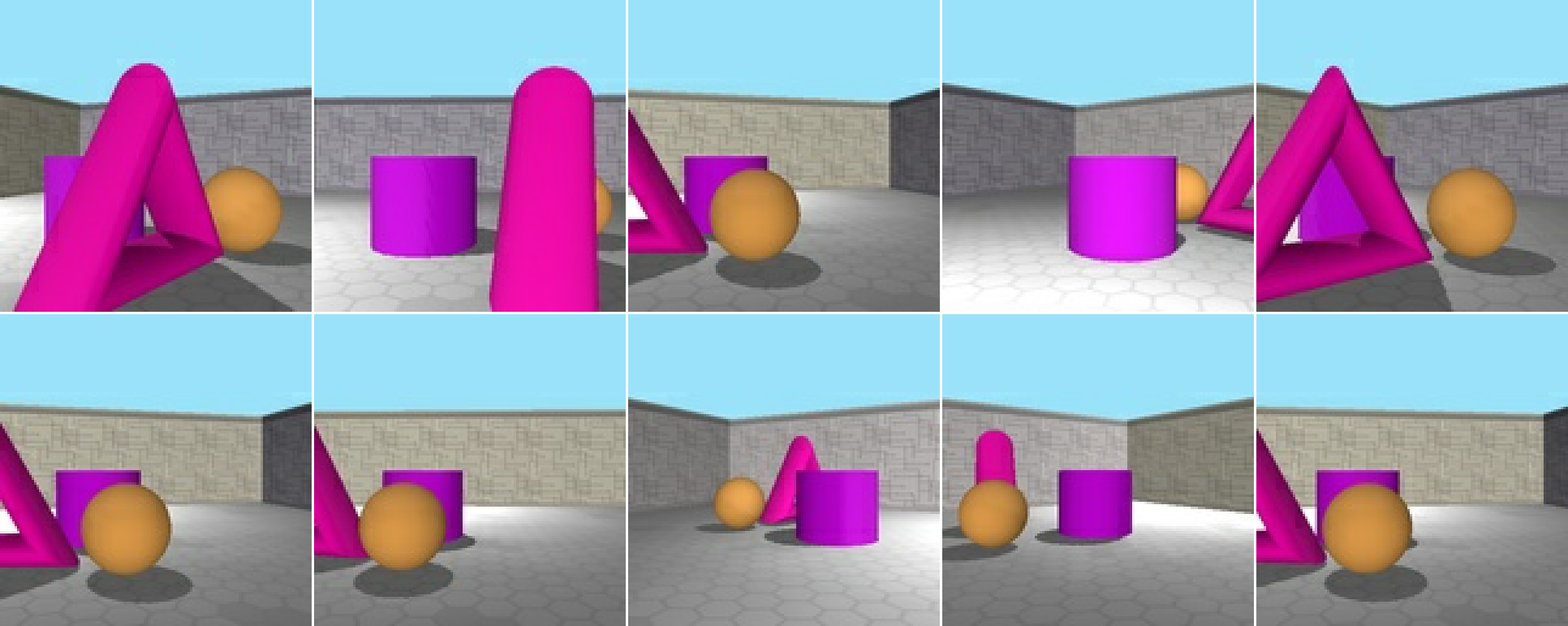} & There is a huge lavender triangle on the left side of the room
and a purple cube behind it on the left side and a light brown ball on
the right side of it.\\
\includegraphics[trim={13.6cm 13.6cm 40.8cm 0},clip, width=2.0cm]{nl_3obj} & In the middle of the left side of the frame is a purple
cylinder shape. Close up on the right side of the screen is a pink
rounded cone almost sphere like at the top. It about an inch taller
than the cylinder. Behind the pink shape is a minor protrusion of an
orange sphere it comes out about halfway down the pink shape and
barely pokes out (.25 in)\\
\includegraphics[trim={27.2cm 13.6cm 27.2cm 0},clip, width=2.0cm]{nl_3obj} & There is a yellow sphere in the middle of the image. Behind the
sphere to the left, is a purple cube the majority of which is covered
by sphere. There is a hot pink three dimensional triangle made of
three dimensional cylinders, with an open center. Half of this
triangle is covered by the left.\\
\includegraphics[trim={40.8cm 13.6cm 13.6cm 0},clip, width=2.0cm]{nl_3obj} & The ceiling is sky blue. The floor has cream-colored, six-sided
tiles throughout. There are two side-half walls the same color as the
floor. There is a neon-purple colored wide cylinder shape in the
center of the floor, behind that and slightly to the right is a yellow
shaped semicircle shape, and to the right of that is a neon pink shape
of part of a triangle slightly showing.\\
\includegraphics[trim={54.4cm 13.6cm 0 0},clip, width=2.0cm]{nl_3obj} & a light grey room has a light blue ceiling. a large pink
triangle is in front left and yellow ball is in the center.\\
\includegraphics[trim={0 0 54.4cm 13.6cm},clip, width=2.0cm]{nl_3obj} & There is a pink triangle to the left of a yellow ball. behind
them is a pink cylinder.\\
\includegraphics[trim={13.6cm 0 40.8cm 13.6cm},clip, width=2.0cm]{nl_3obj} & an oranger sphere sits in grey room next to a pink triangle and
pink cylinder.\\
\includegraphics[trim={27.2cm 0 27.2cm 13.6cm},clip, width=2.0cm]{nl_3obj} & It's a grey brick room and floor with three geometric shapes in
the center. There is a purple cylinder, pink triangle behind that, and
a yellow ball in the rear.\\
\includegraphics[trim={40.8cm 0 13.6cm 13.6cm},clip, width=2.0cm]{nl_3obj} & pink rounded at the top cylinder object in front of it is a
orange ball to the right is a purple big round cylinder slightly
taller then the ball. baby blue wall on top grey boarder horizontal
then light grey under boarder. the floor is grey pattern octagon
shaped.\\
\includegraphics[trim={54.4cm 0 0 13.6cm},clip, width=2.0cm]{nl_3obj} & In the left front center of the screen is a tan ball with a
flat dome shape peeking from behind and a giant hollow purple triangle
half visible on the left of the screen.\\
\bottomrule
\end{tabular}

\end{document}